\documentclass[lettersize,journal]{IEEEtran}

\usepackage{xspace}
\usepackage{enumitem}
\usepackage[utf8]{inputenc} %
\usepackage[T1]{fontenc}    %
\usepackage{hyperref}       %
\usepackage{url}            %
\usepackage{booktabs}       %
\usepackage{amsfonts}       %
\usepackage{nicefrac}       %
\usepackage{microtype}      %
\usepackage[table,svgnames,x11names]{xcolor}         %

\usepackage{amsmath}
\usepackage{amssymb}
\usepackage{mathtools}
\usepackage{amsthm}

\usepackage[linesnumbered,ruled,vlined]{algorithm2e}

\SetCommentSty{mycommfont}
\SetKwInput{KwInput}{Input}                %
\SetKwInput{KwOutput}{Output}              %
\SetKwInput{KwRequire}{Require}              %
\usepackage{array}
\usepackage{soul}
\usepackage{multirow}
\usepackage{xtab,rotating}
\usepackage{subfig}
\usepackage[export]{adjustbox}
\usepackage{wrapfig}

\definecolor{stelios_colour}{RGB}{200, 238, 200}

\newcolumntype{s}{>{\columncolor{Lavender}} c}
\newcolumntype{d}{>{\columncolor{Thistle3}} c}
\newcolumntype{f}{>{\columncolor{LightPink1}} c}

\newcommand{\sysname}{\textit{MetaCLBench}\xspace}

\begin{document}

\title{MetaCLBench: Meta Continual Learning Benchmark on Resource-Constrained Edge Devices}

\author{Sijia Li*, Young D. Kwon*, Lik-Hang Lee~\IEEEmembership{IEEE Senior Member}, and Pan Hui~\IEEEmembership{IEEE Fellow}
\thanks{*Co-first authors. Sijia Li and Pan Hui are affiliated with Hong Kong University of Science and Technology (Guangzhou); Young D. Kwon is affiliated with Samsung AI Center-Cambridge; Lik-Hang Lee is affiliated with The Hong Kong Polytechnic University.}%
\thanks{\textit{Corresponding author: Young D. Kwon} (email: yd.kwon@samsung.com)}
}

\maketitle

\vspace{-2mm}
\begin{abstract}
  Meta-Continual Learning (Meta-CL) enables models to learn new classes from limited labelled samples, making it promising for IoT applications where manual labelling is costly. However, existing studies focus on accuracy while ignoring deployment viability on resource-constrained hardware. Thus, we present \sysname, a benchmark framework that evaluates Meta-CL methods for both accuracy and deployment-critical metrics (memory footprint, latency, and energy consumption) on real IoT devices with RAM sizes ranging from 512\,MB to 4\,GB.
  We evaluate six Meta-CL methods across three architectures (CNN, YAMNet, ViT) and five datasets spanning image and audio modalities. 
  Our evaluation reveals that, depending on the dataset, up to three of six methods cause out-of-memory failures on sub-1 GB devices, significantly narrowing viable deployment options.
  LifeLearner achieves near-oracle accuracy while consuming 2.54-7.43$\times$ less energy than the Oracle method.
  Notably, larger or more sophisticated architectures such as ViT and YAMNet do not necessarily yield better Meta-CL performance, with results varying across datasets and modalities, challenging conventional assumptions about model complexity.
  Finally, we provide practical deployment guidelines and will release our framework upon publication to enable fair evaluation across both accuracy and system-level metrics.
\end{abstract}

\begin{IEEEkeywords}
Continual Learning, Meta-Learning, Few-Shot Learning, IoT, Deployment, On-device Learning
\end{IEEEkeywords}

\vspace{-1mm}
\section{Introduction}\label{sec:introduction}

IoT sensing applications~\cite{iotj_remote_sensing_1_haq,iotj_remote_sensing_2_karim,iotj_realtimeml_survey,jia2023ur2m} increasingly require models that adapt post-deployment~\cite{jia2024tinytta}. Smart speakers must recognise new voice commands~\cite{michieli23_interspeech} as users' vocabularies evolve~\cite{di2024patterns}; industrial acoustic monitors must identify novel machinery sounds and/or anomalies~\cite {banbury2021micronets} indicating emerging faults~\cite{banbury2021mlperf}; object detectors deployed in the wild must accommodate new object categories as environments change~\cite{MAI202228}. These scenarios share two constraints: models operate on resource-limited hardware (typically under 1 GB RAM)~\cite{kwon_yono_ipsn22}, and labelled training data for new classes is scarce~\cite{kwon2023lifelearner,kwon2024tinytrain}. A fundamental obstacle arises when models learn new classes: Catastrophic Forgetting (CF)~\cite{MCCLOSKEY1989109}, where acquiring new knowledge degrades performance on previously learned classes. Also, addressing CF on edge devices is challenging as solutions must fit within tight memory and energy budgets.

Continual Learning (CL)~\cite{shin2017continual,chaudhry2019continual} addresses CF by enabling models to learn sequential tasks while retaining prior knowledge. Existing approaches are grouped into three categories: (1) \textit{rehearsal-based methods} that store and replay past samples~\cite{chauhan2020contauth,pellegrini2020latent,rebuffi2017icarl}, (2) \textit{regularisation-based methods} that constrain weight updates to preserve important parameters~\cite{kirkpatrick2017overcoming,zenke2017continual}, and (3) \textit{parameter isolation methods} that allocate separate model capacity for each task~\cite{hung2019compacting,rusu2016progressive,yoon2017lifelong}. Rehearsal-based approaches achieve the strongest accuracy but require substantial labelled data~\cite{parisi2019continual} and impose significant memory and computational overheads~\cite{kwon_exploring_sec21}, often exceeding what typical IoT devices can accommodate.

To address these limitations, \textit{Meta-Continual Learning (Meta-CL)} has been introduced by integrating few-shot learning with CL~\cite{javed2019meta,jerfel2019reconciling}. Through meta-training on source tasks, Meta-CL methods learn representations that enable rapid adaptation to new classes from only a few labelled samples to minimize the burden of manual labeling of conventional CL methods~\cite{javed2019meta,beaulieu2020learning,lee2021few} and optimize resource utilization through algorithm-system co-design~\cite{kwon2023lifelearner}. 
Despite the benefits of Meta-CL, there are still notable limitations that warrant further investigation.
\textbf{First}, many existing studies~\cite{javed2019meta,beaulieu2020learning,lee2021few} primarily focus on accuracy, neglecting deployment viability on resource-constrained hardware (e.g., A method achieving high accuracy is unusable if it requires 2 GB RAM on a device with only 512 MB available.)
\textbf{Second}, the effectiveness of the existing methods, which are typically used for tasks involving images, has not been investigated fully with sequential time series data provided by auditory sensor systems, where the modality of the data is significantly different from images~\cite{purwins_deep_2019}.
\textbf{Lastly}, the datasets tested have been relatively simple and limited in their variety~\cite{kwon2023lifelearner}, consisting of a few image/audio or specially crafted datasets that lack generalizability.

In this paper, we address these gaps through a systematic benchmark study of six representative Meta-CL methods across three network architectures and five datasets on real IoT hardware. The employed methods include regularisation-based approaches ((1) OML~\cite{javed2019meta}, (2) ANML~\cite{beaulieu2020learning}, (3) OML+AIM, and (4) ANML+AIM~\cite{lee2021few}) and rehearsal-based approaches ((5) Latent OML and (6) LifeLearner~\cite{kwon2023lifelearner}).

Furthermore, prior works of Meta-CL focus on image classification, leaving audio sensing applications unexplored despite their importance for IoT (to name a few applications: voice interfaces, acoustic monitoring). We therefore include audio datasets for keyword spotting (GSCv2~\cite{warden2018speech}) and environmental sound classification (UrbanSound8K~\cite{salamon2014dataset}, ESC-50~\cite{piczak2015esc}), alongside image datasets (CIFAR-100~\cite{krizhevsky2009learning}, MiniImageNet~\cite{vinyals2016matching}). In addition, prior studies~\cite{beaulieu2020learning,lee2021few,kwon2023lifelearner} evaluate only three-layer CNN architectures. We extend the evaluation to 
YAMNet~\cite{howard2017mobilenets}, a MobileNet-based architecture for 
audio, and Vision Transformers (ViT)~\cite{dosovitskiy2020image} for image to conduct a comprehensive benchmark study.

Unlike prior benchmarks reporting only accuracy on server-class machines, we evaluate on real IoT devices, including Jetson Nano (4 GB), Raspberry Pi 3B+ (1 GB), and Pi Zero 2 (512 MB), measuring peak memory footprint, end-to-end latency, and energy consumption alongside accuracy. This enables practitioners to assess deployment viability, not just model quality. Overall, our key contributions include:

\begin{itemize}
    \item \textbf{Edge-Deployable Benchmark Framework (Section~\ref{sec:benchmark}):} 
    We develop \textbf{\sysname} for evaluating on-device Meta-CL across IoT sensing modalities. The framework incorporates audio datasets for smart home and industrial applications (GSCv2, UrbanSound8K, ESC-50) and visual datasets (CIFAR-100, MiniImageNet), measuring deployment-critical metrics (memory footprint, latency, and energy) on real IoT hardware from 512 MB to 4 GB device classes.
    \item \textbf{Deployment-Oriented Findings (Section~\ref{sec:evaluation}):} 
    We reveal that, depending on the dataset, up to three of six methods cause out-of-memory (OOM) issues on sub-1 GB devices (see Table~\ref{tab:accuracy} for ESC-50 results), substantially narrowing viable IoT deployment options. 
    Our analysis quantifies the computational and memory demands on edge devices (Section~\ref{subsec:performance})
    We quantify energy-accuracy trade-offs: LifeLearner achieves near-oracle accuracy at 2.54-7.43$\times$ lower energy consumption than the Oracle method.
    Notably, we find that well-tuned basic 3-layer CNN architecture with proper pre-training can outperform more sophisticated models such as ViT and YAMNet (Tables~\ref{tab:accuracy} and~\ref{tab:accuracy_2}) under difficult Meta-CL setups, challenging conventional assumptions about model complexity.
    \item \textbf{Open Tools and Practical Guidelines (Section~\ref{sec:guideline}):} 
    By implementing \sysname on IoT devices, we provide realistic insights and concrete deployment guidelines specifically tailored for resource-constrained devices.
    Furthermore, we plan to release our benchmark framework and evaluation tools to enable fair and comprehensive assessment of both accuracy and system-level performance metrics. This will facilitate reproducible research and accelerate the development of practical Meta-CL solutions for real-world IoT applications.
\end{itemize}

\vspace{-1mm}
\section{Related Work}\label{sec:related work}

\textbf{Continual Learning:} CL aims to learn new knowledge from dynamically evolving input data distributions~\cite{MAI202228,CL_survey_2024}. When algorithms are trained sequentially, the new learning will interfere with previous established weights, leading to CF of the previous learning~\cite{MCCLOSKEY1989109}. Current CL approaches to the CF challenge can be divided into two categories based on task information management. The first approach incorporates regularization terms into the loss function to facilitate knowledge of old and new models, thereby preserving previously learned knowledge~\cite{Rannen_2017_ICCV,8107520,DBLP872,ahn2019uncertainty}. The second approach involves either original or pseudo-samples from previous tasks during new task training~\cite{shin2017continual,rolnick2019experience,chaudhry2019continual}.

\textbf{Meta Continual Learning:} Meta-CL updates the model in the outer loop using learned random samples and optimizes it in the inner loop with a few samples before passing it to the outer loop~\cite{jerfel2019reconciling}. Previous research on Meta-CL has primarily concentrated on image recognition applications. Online-aware Meta-Learning (OML)~\cite{javed2019meta} and A Neuromodulated Meta-Learning (ANML)~\cite{beaulieu2020learning} both demonstrated high CL performance on the Omniglot dataset~\cite{lake2015human}. Based on these approaches, state-of-the-art (SOTA) Meta-CL incorporated the Attentive Independent Mechanisms (AIM) module, which can independently acquire new knowledge to enhance model performance~\cite{lee2021few}. However, due to its high processing cost, using Meta-CL is challenging on low-end edge devices~\cite{kwon2023lifelearner}. LifeLearner addresses these limitations by integrating meta-learning, rehearsal mechanisms, and resource-efficient compression, making it suitable for deploying Meta-CL applications on resource-constrained devices~\cite{kwon2023lifelearner}.

\textbf{Continual Learning for Image and Audio Applications:} CL has remarkable capabilities in sensing applications, including action recognition, audio sensing, and speech recognition~\cite{jha_continual_2021,kwon_exploring_sec21}. Its ability to update models based on continuous data streams makes it particularly valuable for mobile devices processing speech and image data. Researchers have successfully applied CL to speech and emotion recognition~\cite{thuseethan2021deep}. CL systems integrated into edge devices with audio and microphones continually learn from ambient human voices and environmental sounds to improve recognition accuracy~\cite{kwon21_interspeech}. In image recognition, CL facilitates applications such as food recognition~\cite{he2021online}, where systems must continuously learn from new instances to enhance recognition accuracy.

Prior work demonstrates CL's value for IoT sensing applications, yet does 
not address a fundamental question facing practitioners: \textit{which methods can actually deploy on target hardware while achieving acceptable accuracy?} Our work fills this gap by evaluating Meta-CL methods for deployment viability, not just accuracy. We measure memory footprint, latency, and energy consumption on three device classes, revealing that several methods optimized for accuracy fail to run on typical IoT hardware. This deployment-oriented evaluation across image and audio modalities, using architectures from lightweight CNNs to ViT and YAMNet, provides actionable guidance for IoT system designers.

\begin{figure}[t]
    \centering
    \vspace{-0.3cm}
    \includegraphics[width=1\columnwidth]{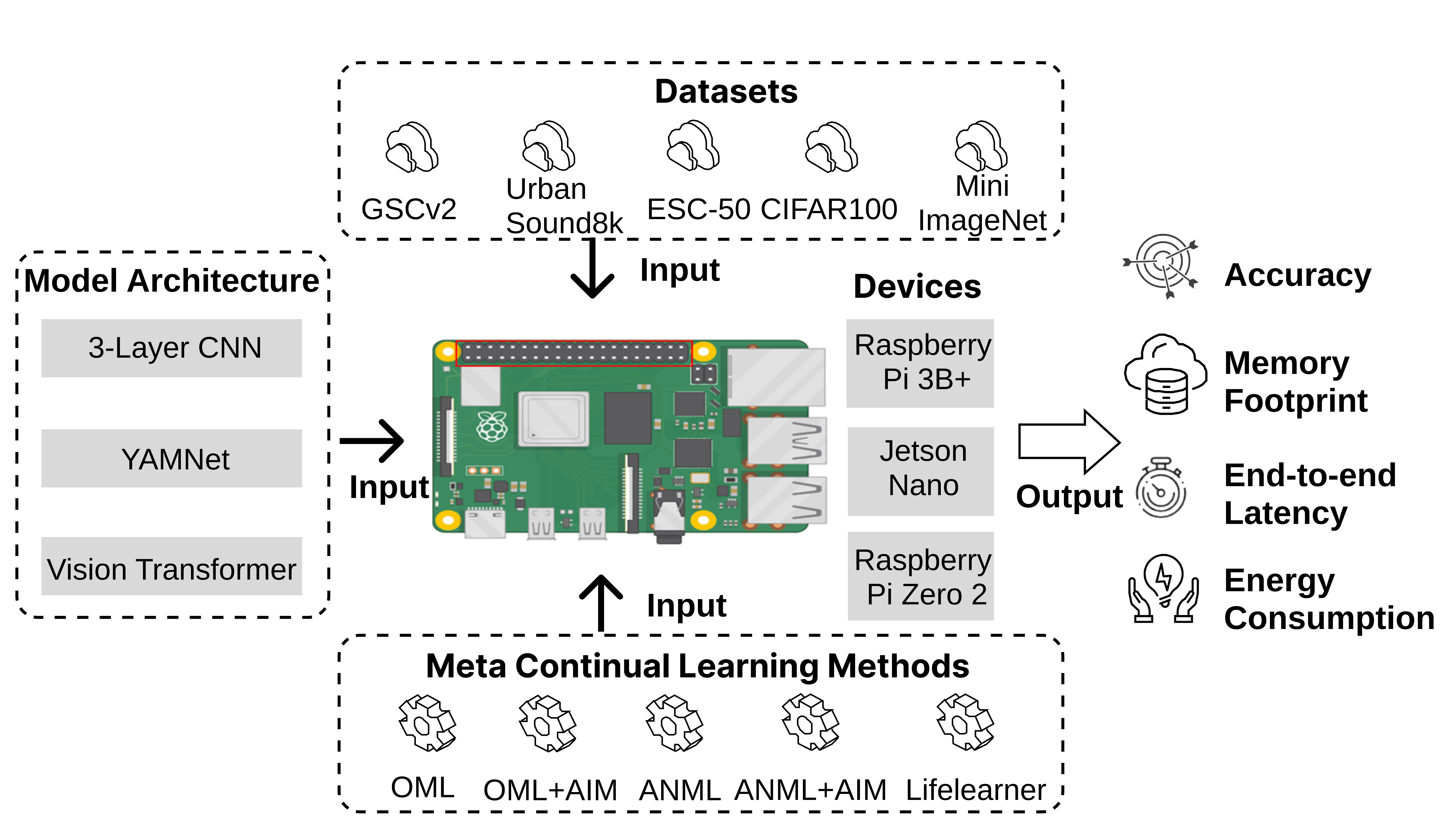}
    \caption{The framework overview. Testing trade-offs between performance and system resources across three devices with five datasets and six Meta-CL methods using three architectures.}
    \label{fig:framework}
    \vspace{-0.4cm}
\end{figure}

\vspace{-1mm}
\section{Benchmark Framework}\label{sec:benchmark}

\subsection{Continual Learning Task}\label{subsec:cl_task}

\textbf{Problem Formulation:} CL addresses the challenge of learning new tasks without forgetting previously learned ones. Given a sequence of tasks $S$, the objective is to maintain robust performance across all tasks $S_{1....n}$. Our evaluation framework encompasses ($T (Task), S (Sequence), D (Datasets)$), where each task represents a distinct sound class. Each subsequent task $Sn$ is derived from the historical training records of the most recent sets $S_{n-1}, S_{n-2}...S_{n-m}$ of the employed audio and image datasets.

Meta-Learning is conducted in both the inner and outer loops, enabling the training of each inner loop round through the outer loop. Each inner loop iteration is performing a complete learning process, evaluating both the acquisition of new tasks and the retention of previously learned ones. The meta-loss gradient  (comprising newly learned errors and errors from random sampling of the dataset) propagates back to update the initial parameters. This process continues iteratively as new tasks are incorporated. By constraining each inner loop to a single new task while continuously undergoing meta-training, it mitigates CF and optimizes computational efficiency.

\textbf{Our Framework Overview:} 
Figure~\ref{fig:framework} shows our experimental framework, \sysname, designed to evaluate the performance and resource trade-offs across multiple dimensions: three devices, five datasets, six Meta-CL methods, and three architectures. \sysname extends Meta-CL by implementing a comprehensive evaluation system that spans multiple modalities and practical considerations. By combining both visual and audio datasets, we provide thorough insights into Meta-CL performance across diverse data types. We conduct a detailed system-level analysis of computational efficiency, memory usage, and energy consumption, particularly on resource-constrained edge devices for real-world applications. Overall, our framework evaluates six established Meta-CL methods across three network architectures of varying complexity, creating a robust benchmark framework that assesses both theoretical capabilities and practical constraints.
We now explain each of the components of our framework in the following subsections.

\vspace{-1mm}
\subsection{Meta-CL Methods}\label{subsec:methods}

\begin{figure}[t]
    \centering
    \vspace{-0.5cm}
    \includegraphics[width=1.04\columnwidth]{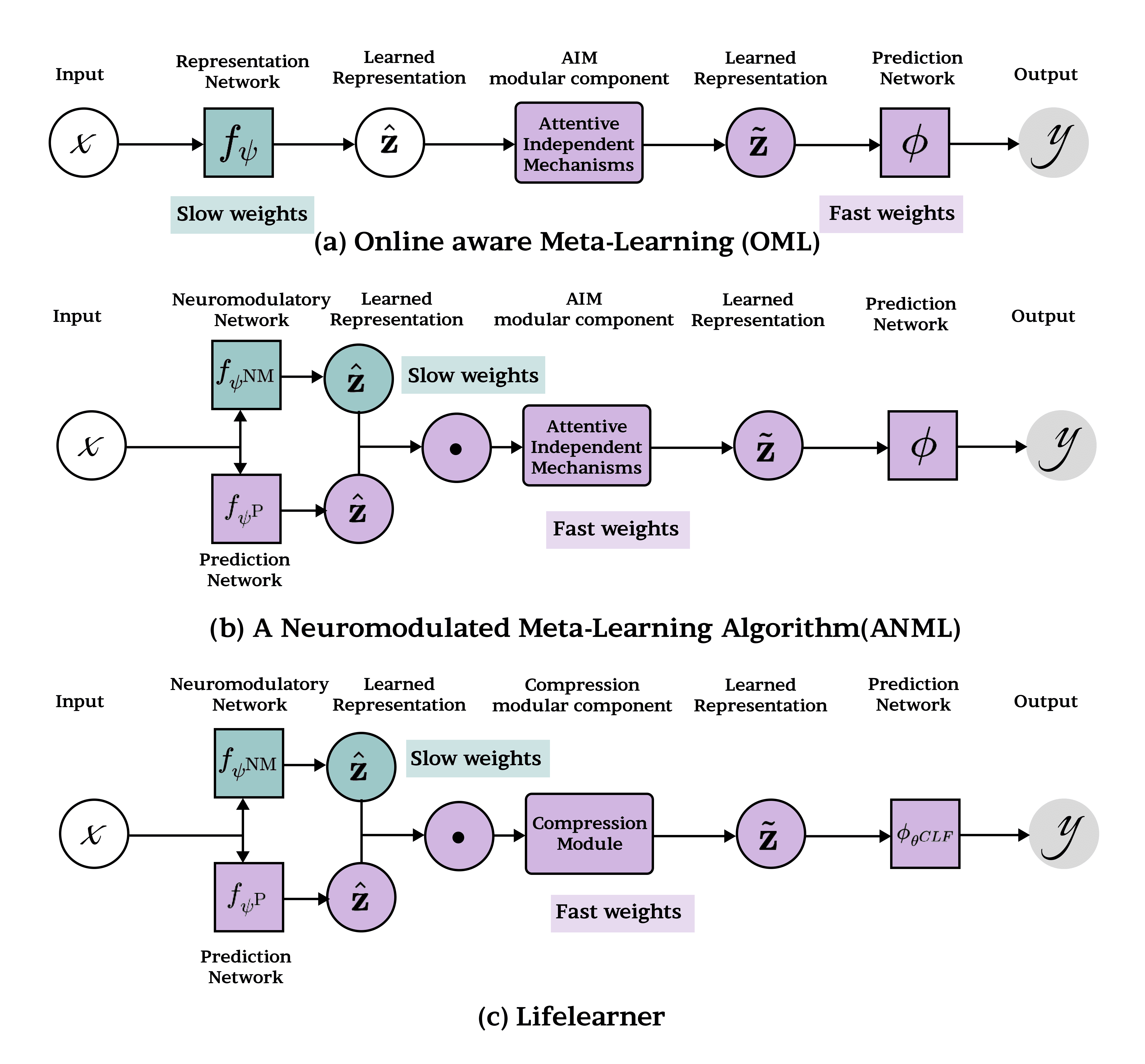}
    \caption{The illustration of the Meta-CL methods evaluated in our benchmark framework.}
    \label{fig:methods}
    \vspace{-0.3cm}
\end{figure}

Figure~\ref{fig:methods} presents six distinct Meta-CL methods: Online aware Meta-Learning (OML), OML with Attentive Independent Mechanisms (OML+AIM), A Neuromodulated Meta-Learning Algorithm (ANML), ANML+AIM, and LifeLearner. The following sections detail each method.

\textbf{Oracle}: This baseline represents the case where it has access to all the classes at once in an i.i.d. fashion. Also, we assume that Oracle would perform DNN training for multiple epochs until the model converges. As a result, Oracle often shows superior accuracy, and thus its accuracy represents the \textit{upper bound} of our evaluation.

\textbf{OML~\cite{javed2019meta}} continuously optimizes representations during online updates to quickly acquire predictions for new tasks, facilitating seamless and efficient CL. This approach uses CF  as signals to justify the use of storage space for accommodating new tasks. The OML can accomplish CL in more than 200 consecutive classes. It is an online updating method that constantly learns representations and often outperforms rehearsal-based continuous learning methods.

\textbf{ANML~\cite{beaulieu2020learning}:} This is an approach inspired by the neuromodulation process in the brain. This method accomplishes selective activation (forward pass) and selective plasticity (backward pass) in DNNs by meta-learning through a CL process within the prediction network ($f_{\varphi P}$). Optimizing activation timing by controlling the activation of a predictive model based on input conditions enhances continual learning. It is often reported to be superior to OML, as it can be optimized to reduce both forward and backward interference.

\textbf{OML+AIM~\cite{lee2021few}:} This is a Meta-CL approach that integrates Attentive Independent Mechanisms (AIM), which is a modular component into the OML framework. AIM increases the speed of learning a new task by selecting out the mechanism that best explains the representation of the inner loop to activate.

\textbf{ANML+AIM~\cite{lee2021few}:} This is a Meta-CL approach that incorporates AIM into the ANML framework. The accuracy of SOTA Meta-CL continues to improve beyond the performance of the previous three approaches by integrating AIM into the existing CL framework.

\textbf{LifeLearner~\cite{kwon2023lifelearner}:} This module is based on ANML and combines both lossy and lossless compression to achieve high compression rates and minimize footprint. The LifeLearner approach not only achieves optimal CL efficiency but also significantly reduces energy consumption and latency compared to the SOTA. And it achieves efficient learning even on resource-constrained edge devices.

\textbf{Latent OML:} This module is based on OML and incorporates the rehearsal strategy from LifeLearner to achieve high Meta-CL accuracy. It is used to evaluate advanced architectures (YAMNet and ViT).

\vspace{-2mm}
\subsection{Datasets}\label{subsec:Dataset}
To complement the relative simplicity, limited variety, and lack of generalizability of the datasets tested in previous studies, we added three audio datasets to the single-label image datasets (\textit{CIFAR-100}~\cite{krizhevsky2009learning}, \textit{MiniImageNet}~\cite{vinyals2016matching}) from previous studies. Specifically, we used the \textit{UrbanSound8K}~\cite{salamon2014dataset} dataset with fewer classes, the \textit{ESC-50}~\cite{piczak2015esc} dataset with more classes, and the \textit{GSCv2}~\cite{warden2018speech} dataset for keyword spotting (KWS). 

\textbf{Google Speech Commands V2 (GSCV2)}~\cite{warden2018speech} is a massive database of one-second voice clips of 30 short words consisting of a solitary spoken English word or ambient noise divided into 35 classes with 105,829 clips. These clips are derived from a limited number of commands and are spoken by various speakers. Each class has 2,424 input data for training and 314 for testing, with 25 classes allocated for meta-training and the remaining 10 for meta-testing. Then, up to 30 samples are provided for each class during meta-testing.

\textbf{UrbanSound8K} is a comprehensive collection for environment sound classification (ESC) applications comprising 8,732 labeled sounds that are no more than four seconds~\cite{salamon2014dataset}. This single-labeled dataset is categorized into ten classes from the taxonomy in real-world settings, including children playing sounds, street music, gunshots, etc. Referring to previous studies~\cite{su2019environment}, indicate that four features %
in each audio clip were extracted and resampled to 22 kHz. Among the ten classes, half of them are allocated for meta-training, while the other half are specifically assigned for meta-testing. Up to 30 samples are given for each class during the meta-testing. Based on the first three seconds of each audio segment, the input to the audio is 128 \textit{(number of frames)} $\times$ 85 \textit{(size of the set of four features)}.

\textbf{ESC-50} is a smaller volume collection consisting of 2,000 environment sounds in 5-second segments~\cite{piczak2015esc}. This single-labeled dataset is categorized into 50 classes and divided into five groups: animals, natural soundscapes, human sound without speech, domestic sounds and urban noises. Out of the 50 classes. 30 are designated for meta-training and the remaining 20 for meta-testing. Then, up to 30 samples are given for each class during meta-testing. As part of the preprocessing procedure, we resampled the ESC-50 samples to 32kHz and generated an input with dimensions of 157 \textit{(number of frames)} $\times$ 64 \textit{(features)}~\cite{chen2022hts,pons2019training}.

\textbf{CIFAR-100} consists of 60,000 images distributed across 100 classes and is widely used for training and assessing various machine learning algorithms in tasks related to image classification~\cite{krizhevsky2009learning}. Each class contains 500 photographs for training and 100 images for testing. Of the 100 classes, 70 are designated for meta-training and 30 for meta-testing. In both phases, 30 random training images per class are utilized. For the meta-testing phase, 900 samples are utilized during meta-testing to carry out CL.

\textbf{MiniImageNet} is excerpted from the Imagenet dataset and is utilized in few-shot learning research~\cite{vinyals2016matching}. Based on the previous research, the MiniImageNet dataset is composed of 64 classes designated for meta-training and 20 for meta-testing. Each class offers 540 images for training and 60 for testing. Moreover, a total of 600 samples are available during the meta-testing phase.

\vspace{-1mm}
\subsection{Model Architecture}
We augmented the 3-layer CNN architecture previously employed in CL studies by incorporating the YAMNet network architecture with pre-training. In addition to assessing YAMNet as a model for the speech domain with pre-training, we also evaluated the ViT, a transformer model that integrates a pre-trained model used for image classification by segmenting the image to a fixed size for input to the transformer encoder. 

\textbf{3-layer Architecture:} The methods depicted in Fig. 2 consist of feature extractors and classifiers. The feature extractor transmits the learned features to the AIM and the compression modules, followed by classification. The $f_\varphi W$ added to the AIM module components can be used to minimize forgetting and learn new categories efficiently. In the 3-layer architecture, the feature extractor for OML and OML+AIM incorporates $f_\varphi$, which is a 6-layer CNN with 112 channels, and the classifier has two fully-connected layers. And the LifeLearner and ANML+AIM share the same natural structure. The feature extractor includes a neuromodulatory network ($f_{\varphi NM}$) and a prediction network ($f_{\varphi P}$), which is a 3-layer CNN with 112 channels. The classifier consists of a single fully-connected layer with 256 channels.

\textbf{YAMNet}~\cite{howard2017mobilenets} employs the MobileNetV1, which features a depthwise separable convolution architecture designed to decrease model size and latency and a full convolution on the first layer. The depthwise separable convolution comprises a deep convolution responsible for filtering and a pointwise convolution used $1\times 1$ convolution responsible for combining features, effectively mixing information from all output channels in the deep convolution step. MobileNetV1 consists of 28 layers, with each layer being downsampled through strided convolution. Additionally, MobileNets introduces two hyperparameters, a width multiplier and a resolution multiplier, which facilitate the balance between latency and accuracy.

\textbf{ViT}~\cite{dosovitskiy2020image} embodies a hybrid architecture. Since ViT only has a Multilayer Perceptron (MLP) layer with localization and translation equivalence, the image-specific generalization bias of ViT is much smaller than that of CNN. The same five methodologies deployed in the 3-layer architecture have been adapted for use with ViT. ViT begins by partitioning an image into patches; these patches, extracted from the CNN feature maps, serve as alternative input sequences for the model.

\vspace{-1mm}
\subsection{System Implementation}
The implementation of our benchmark framework, \sysname, consists of two stages. We developed the first stage, pre-training and meta-training, of Meta CL methods on a Linux server to initialize and optimize neural weights in order to enable fast adaptation during deployment scenarios with a few samples.
After that, we implemented the second stage, meta-testing (i.e., actual deployment setup), on our target devices: (1) embedded and mobile systems such as Jetson Nano and Raspberry Pi 3B+, and (2) a severely resource-constrained IoT device like Raspberry Pi Zero 2. 
In addition, for methods that utilize hardware-friendly optimization (e.g., LifeLearner), we also adopt their optimization techniques in our framework. 
Specifically, \sysname incorporates hardware-friendly 8-bit integer arithmetic~\cite{jacob_quantization_2018} which reduces the precision of weights/activations of the model from 32-bit floats to 8-bit integers, increasing the computation throughput and minimizing latency and energy. \sysname uses the scalar quantization scheme~\cite{jacob_quantization_2018} to minimize the information loss in quantization. Also, the QNNPACK backend engine is used to execute the quantized model on the embedded devices.

\textbf{Target IoT Hardware:} We evaluate on three device classes 
representing the IoT hardware spectrum:

\begin{itemize}[leftmargin=*]
    \item Jetson Nano (4\,GB): Entry-level edge AI platform, representative of smart cameras and local inference gateways.
    \item Raspberry Pi 3B+ (1\,GB): General-purpose IoT device, representative of smart home hubs and sensor gateways.
    \item Raspberry Pi Zero 2 (512\,MB): Severely constrained platform, representative of wearables and low-cost distributed sensors.
\end{itemize}

The gap between nominal and available RAM---caused by OS overhead and 
background services---is critical for deployment planning but often 
overlooked in ML benchmarks. Available memory during item time is 
approximately 1.7\,GB, 600\,MB, and 250\,MB respectively. Hence, we allocate an additional 1GB of swap space, enabling the execution of memory-intensive Meta-CL methods such as ANML+AIM and OML+AIM. Our software stack includes Faiss~\cite{johnson_billion-scale_2019} and PyTorch 1.13 for meta-training and meta-testing stages.

\vspace{-1mm}
\subsection{Experimental Setup}\label{subsec:Experimental Setup}
\vspace{-1mm}

\textbf{Training Detail:}
We adopt a pre-training procedure, similar to prior works~\cite{hu2022pushing,yosinski2014transferable}. For the advanced model architectures such as ViT and YAMNet, there exist pre-trained model weights, thus we utilize them directly similar to~\cite{hu2022pushing}. Then, for the CNN architecture, we pre-train the CNN model with a sufficient number of epochs (i.e., until the validation loss converges), which is consistent with conventional transfer learning for DNNs~\cite{yosinski2014transferable}.
Consistent with prior meta-training research~\cite{lee2021few,kwon2023lifelearner}, we employed a batch size of 1 for the inner-loop updates and 64 for the outer-loop updates across 20,000 steps to ensure the accuracy of our results. To obtain a meta-trained model with the highest possible validation set accuracy, we tested with various learning rates required for both the inner and outer loops. Consequently, we established the learning rate for the inner loop ($\alpha$) was set to 0.001, and the learning rate for the outer loop ($\beta$) was also set to 0.001 for all datasets. In the meta-testing phase, we assessed ten distinct learning rates for each method and reported the best accuracy. To evaluate the precision of rehearsal-based methods, we experimented with batch sizes of 8 and 16 and observed a minimal performance difference depending on the batch sizes. We selected a batch size of 8 as it requires less memory than larger batch sizes.

We chose to use ANML with CNN architectures and OML with more advanced models like ViT and YAMNet. Our preliminary tests revealed minimal differences between ANML and OML within the same architecture, with ANML slightly outperforming OML in terms of accuracy. Due to this marginal difference, we focused on ANML for CNNs to streamline the analysis. For more complex architectures like ViT and YAMNet, we opted for OML, as ANML introduces additional computational overhead due to an extra layer. Our study concentrated on the performance of methods within specific architectures rather than comparing ANML and OML across all architectures. This streamlined approach allowed us to present a clearer analysis without redundant testing across combinations.

\textbf{Evaluation Metrics:}
Following~\cite{beaulieu2020learning}, this study focuses on the accuracy of unseen samples across new categories in CL. We measure the end-to-end latency, energy consumption, and peak memory by deploying various Meta-CL methods using \sysname on edge devices: Jetson Nano, Raspberry Pi 3B+, Raspberry Pi Zero 2.
Specifically, we measure system performance metrics by continually learning all given classes for deployed DNNs on embedded devices. Peak memory includes: (1) model memory for updated weights, (2) optimizer memory for gradients and momentum, (3) activations memory for intermediate outputs during weight updates, and (4) rehearsal samples. To measure end-to-end latency and energy consumption, we account for the time and energy required to: (1) load the model and (2) perform Meta-CL using all available samples (\textit{e.g.,} 30) across all classes (\textit{e.g.,} 30 for CIFAR-100 and 10 for GSCv2). Energy consumption is measured by monitoring the power consumption of the edge device using a YOTINO USB power meter. We derive the energy consumption using the equation (Energy = Power x Time).

\vspace{-1mm}
\section{Results \& Findings}\label{sec:evaluation}

This section presents comprehensive evaluation results and findings from \sysname.

\vspace{-2mm}
\subsection{Main Results}\label{subsec:performance}

We now present deployment-oriented findings, focusing on both accuracy and deployment-critical metrics that determine whether methods are viable for IoT deployment. As previous research~\cite{beaulieu2020learning,holla2020meta} has shown that ANML often outperforms OML when using 3-layer CNN architectures, we evaluate ANML on CNN architectures. For advanced models like YAMNet and ViT, we prioritized OML evaluation due to their inherent pre-training components, which eliminate the need for additional pre-training through ANML or the AIM module—additions that neither significantly improve accuracy nor justify their computational cost.

\textbf{Accuracy.}
Tables~\ref{tab:accuracy} and~\ref{tab:accuracy_2} present accuracy across all method-dataset combinations. Oracle models (Oracle ANML for CNN, Oracle OML for YAMNet and ViT), trained on all classes simultaneously, establish upper-bound accuracy representing ideal (but unrealistic) deployment scenarios.
The key question for IoT practitioners is not which method achieves the highest accuracy, but which methods achieve acceptable accuracy while remaining deployable on target hardware.

By comparing Meta-CL methods against Oracle models, we demonstrate the trade-offs between Meta-CL approaches and ideal scenarios where all data is available initially, highlighting the methods' efficiency under practical constraints.
Also, "Pre-trained" approach (transfer learning applied on pre-trained weights without meta-training) shows lower accuracy, averaging 22.18\% and 13.8\% across the datasets for Convolutional Neural Networks (CNNs) and advanced architectures (YAMNet and ViT). This result indicates that traditional transfer learning fails to address few-shot learning challenges. In CNN architecture, baseline ANML improved upon the "pre-trained" approach with a 16.7\% average accuracy gain. The AIM enhancement further increased accuracy by 2.54\%. However, these improvements remained below Oracle and LifeLearner by 15.92\%.

\textbf{Peak Memory Footprint.}
For IoT deployment, peak memory during learning---not inference---determines viability. Hence, we conducted measurements of the maximum memory requirements for backpropagation and rehearsal samples.
The backpropagation memory comprises: (1) model parameters; (2) optimizer states (gradients and momentum vectors); (3) activation (intermediate tensors used during backpropagation).

First, the results in Tables~\ref{tab:accuracy} and~\ref{tab:accuracy_2} reveal a critical deployment barrier: AIM-enhanced methods (ANML+AIM, OML+AIM) requires 608-2,648\,MB (except for a single case), exceeding available RAM on Pi 3B+ and Pi Zero 2, due to additional parameters and activations required during training AIMs. These methods, despite strong accuracy, are \textit{non-deployable} on typical IoT hardware.

Second, while OML, ANML, Oracle, Pre-trained, and LifeLearner demonstrate lower memory requirements, Pre-trained and ANML are less accurate. LifeLearner emerges as the optimal solution, requiring only 136-496 MB while maintaining accuracy comparable to Oracle. This efficiency is achieved through its compression module for latent replays, reducing activation memory and overall memory footprint, making it particularly suitable for resource-constrained devices.

\textbf{Deployment Viability.} Out-of-memory (OOM) failures occurred for ANML+AIM and OML+AIM on all datasets except GSCv2 and CIFAR-100 when deployed on Pi 3B+. This finding has immediate practical implications: IoT practitioners targeting sub-1\,GB devices should exclude AIM-enhanced methods from consideration regardless of their reported accuracy.

\begin{table}[t]
  \vspace{-0.2cm}
  \centering
  \caption{Performance and computational costs, memory footprint of six representative Meta-CL methods using CNN on five datasets of image and audio domains. OOM indicates an out-of-memory issue.}
  \label{tab:accuracy}
  \resizebox{\columnwidth}{!}{%
  \begin{tabular}{ p {1.4cm} l | c c c c }
    \toprule 
     &  & \multicolumn{4}{c}{CNN} \\ 
    \cmidrule(l){3-6}
     \textbf{Dataset} & \textbf{Method} & \textbf{Accuracy} & \textbf{Memory} & \textbf{Latency} & \textbf{Energy} \\
        \cmidrule(l){0-5}
    \multirow{7}{*}{CIFIAR-100}
    & Pre-trained             & \textbf{0.260}	& 39.69 MB	& \textbf{305s}	& \textbf{1.44 kJ}	\\
    & ANML              & 0.272 & 39.69 MB & 309s & 1.44 kJ  \\
    & ANML+AIM        & 0.346	& \textbf{1,093 MB}	& 6,390s	& 29.43 kJ\\
    &OML+AIM     & 0.311	& 834.1 MB	& 1,481s & 6.84 kJ \\
    & Raw ANML        & 0.392 & 99.5 MB & \textbf{11,424s} & \textbf{52.5 kJ}   \\
    & Oracle ANML        & 0.445 & 39.93 MB & 1,866s & 8.55 kJ  \\  
     & LifeLearner        & \textbf{0.452} & \textbf{15.45 MB} & 374s & 1.71kJ \\
        \cmidrule(l){0-5}
    \multirow{7}{*}{ \hspace{-0.09cm}\parbox{1.4cm}{Mini \\ ImageNet}} & Pre-trained                & 0.258 & 474.5 MB & 1,198s & 5.5 kJ\\
    & ANML                & 0.327 & 474.5 MB & \textbf{1,152s} & \textbf{5.3 kJ} \\
     & ANML+AIM           & 0.331 & \textbf{1,562 MB}& \textbf{OOM} & \textbf{OOM} \\
     & OML+AIM     & \textbf{0.187} &1,051 MB & 1,434s & 6.5 kJ  \\
    & Raw ANML        & 0.429 & 897.1 MB & 185,610s & 810 kJ \\
    & Oracle ANML        & \textbf{0.438} & 475.0 MB & 3,414s & 15.7 kJ \\
    & LifeLearner        & 0.433 & \textbf{137 MB} & 1,343s & 6.17 kJ \\
        \cmidrule(l){0-5}
    \multirow{7}{*}{GSCv2} & Pre-trained                & 0.213 & 10.16 MB & \textbf{71.1s} & \textbf{0.324 kJ}\\
    & ANML                & 0.429 & 10.16 MB & 71.1s & 0.324 kJ \\
    & ANML+AIM           & 0.710 & \textbf{608.2 MB} & 394.2s & 1.8 kJ \\
     & OML+AIM     & 0.649 & 135.2 MB & 157.5s & 0.72 kJ  \\
    & Raw ANML        & \textbf{0.135} & 45.72 MB & \textbf{735.6s} & \textbf{3.37 kJ} \\
    & Oracle ANML        & 0.712 & 10.20 MB & 569.7s & 2.6 kJ \\
    & LifeLearner        & \textbf{0.713} & \textbf{3.40 MB} & 75.6s & 0.35 kJ \\
        \cmidrule(l){0-5}
    \multirow{7}{*}{ \hspace{-0.09cm}\parbox{1.4cm}{Urban \\ Sound8K}} & Pre-trained                & \textbf{0.182}& 1,382 MB & \textbf{302s} & \textbf{1.6 kJ}\\
    & ANML                & 0.596 & 1,382 MB & 305s & 1.6 kJ \\
     & ANML+AIM           & 0.439 & 2,593 MB & \textbf{OOM} & \textbf{OOM} \\
     & OML+AIM     & 0.385 & \textbf{2,648 MB} & OOM & OOM  \\
    & Raw ANML        & 0.667 & 1,456 MB & 60,120s & 279.0 kJ \\
    & Oracle ANML        & \textbf{0.710} & 1,384 MB & 916s & 4.5 kJ \\
    & LifeLearner        & 0.650 & \textbf{496 MB} & 320s & 1.7 kJ \\
        \cmidrule(l){0-5}
    \multirow{7}{*}{ESC-50} & Pre-trained                & 0.196 & 1,163 MB & \textbf{1,359s} & \textbf{7.2 kJ}\\
    & ANML                & 0.308 & 1,163 MB & 1,274s & 7.2 kJ \\
    & ANML+AIM           & 0.233 & 2,305 MB & OOM & OOM \\
    & OML+AIM     & \textbf{0.181} & 2,152 MB & OOM & OOM  \\
    & Raw ANML        & 0.358 & \textbf{5,005 MB} & \textbf{OOM} & \textbf{OOM} \\
    & Oracle ANML       & \textbf{0.435} & 1,167 MB & 4,120s & 20.3 kJ \\
    & LifeLearner        & 0.381 & \textbf{316.3 MB} & 1,445s & 7.8 kJ \\
    \bottomrule
  \end{tabular}
  }
  \vspace{-0.2cm}
\end{table}

\begin{table}[t]
  \centering
  \caption{Performance and computational costs, memory footprint of six representative Meta-CL methods using ViT and YAMNet on five datasets of image and audio domains.}
  
  \label{tab:accuracy_2}
  \resizebox{\columnwidth}{!}{%
  \begin{tabular}{ p {1.4cm} l | c c c c  }
    \toprule 
     &  & \multicolumn{4}{c}{ViT} \\ 
    \cmidrule(l){3-6}
     \textbf{Dataset} & \textbf{Method} & \textbf{Accuracy} & \textbf{Memory} & \textbf{Latency} & \textbf{Energy}  \\
        \cmidrule(l){0-5}
    \multirow{6}{*}
    & Pre-trained             	& \textbf{0.119} & \textbf{29.8 MB} & \textbf{615s} & \textbf{2.96 kJ} \\
     &OML           & 0.271 & 29.8 MB & 628s & 3.03 kJ \\
    CIFAR-100& Raw OML        & \textbf{0.40} & \textbf{153 MB} & \textbf{22,968s} & \textbf{103 kJ}  \\
    & Oracle OML        & 0.361 & 29.5 MB & 3,672s & 19.4 kJ \\  
    & Latent OML         & 0.365 & 36.8 MB & 764s & 3.76 kJ \\
        \cmidrule(l){0-5}
    \multirow{5}{*}{ \hspace{-0.09cm}\parbox{1.4cm}{Mini \\ ImageNet}} & Pre-trained             	& \textbf{0.234} & 337 MB & \textbf{2,056s} & \textbf{11.2 kJ} \\
    &OML           & 0.255 & 337 MB & 2,203s & 11.4 kJ \\
    & Raw OML        & 0.454 & \textbf{1,334 MB} & \textbf{301,456s} & \textbf{1150.0 kJ}  \\
    & Oracle OML        & \textbf{0.512} & 338 MB & 6,301s & 31.0 kJ \\  
    & Latent OML         & 0.376 & \textbf{337 MB} & 2,250s & 13.0 kJ \\
       \cmidrule(l){0-5}
    &  & \multicolumn{4}{c}{YAMNet} \\ 
        \cmidrule(l){3-6}
    & Pre-trained                & \textbf{0.061} & \textbf{39.2 MB} & 213.6s & 1.08 kJ \\
     & OML   & 0.329 & 39.2 MB & \textbf{209.8s} & 1.08 kJ  \\
    GSCv2& Raw OML         & \textbf{0.749} & \textbf{120 MB} & 1,482s & 6.85 kJ \\
    & Oracle OML        & 0.701 & 39.7 MB & \textbf{1,685s} & \textbf{14.6 kJ} \\
    & Latent OML        & 0.704 & 40.9 MB & 235.8s & \textbf{0.43 kJ }\\
        \cmidrule(l){0-5}
    \multirow{5}{*}{ \hspace{-0.09cm}\parbox{1.4cm}{Urban \\ Sound8K}} & Pre-trained             	& \textbf{0.186} & 39.2 MB & 912s & 5.4 kJ \\
     &OML           & 0.262 & 39.2 MB & \textbf{910s} & \textbf{5.36 kJ} \\
     & Raw OML        & 0.595 & \textbf{140 MB} & \textbf{120,965s} & \textbf{550 kJ}  \\
    & Oracle OML        & \textbf{0.729} & 42.4 MB & 2.501s & 21.1 kJ \\  
    & Latent OML         & 0.442 & \textbf{39.2 MB} & 910s & \textbf{1.6 kJ} \\
        \cmidrule(l){0-5}
    & Pre-trained             	& \textbf{0.090} & \textbf{39.8 MB} & 4,052s & 19.6 kJ \\
     &OML           & 0.135 & 39.8 MB & \textbf{3,986s} & 9.6 kJ \\
    ESC-50 & Raw OML        & \textbf{0.577} & \textbf{210.8 MB} & \textbf{28,465s} & 39.6 kJ  \\
    & Oracle OML        & 0.442 & 42.5 MB & 12,473s & \textbf{84.2 kJ} \\  
    & Latent OML         & 0.458 & 44.3 MB & 4,536s & \textbf{7.4 kJ} \\
    \bottomrule
  \end{tabular}
  }
\end{table}

\textbf{End-to-end Latency \& Energy.}
Tables~\ref{tab:accuracy} and~\ref{tab:accuracy_2} also present the operational efficiency of Meta-CL methods on Raspberry Pi 3B+, focusing on end-to-end latency and energy consumption.
The end-to-end latency comprises four components: model loading (5\%), inference (30\%), backpropagation (61\%), and data processing overhead (4\%). 

First, we measured the end-to-end latency of CL methods on the Raspberry Pi 3B+ across all classes (30 samples per class). LifeLearner achieved fast end-to-end latency. Oracle exhibited higher training latency due to i.i.d. training over multiple epochs until it converges. AIM's implementation of attention mechanism selection during backpropagation increased computational overhead. LifeLearner reduces training and memory latency through latent replay and model freezing.
Due to the large amount of memory required by the AIM series, crashes occurred several times during the test due to memory exhaustion. 
In particular, Raw ANML requires substantial latency and energy consumption (e.g., 11,424s and 52.5 kJ on CIFAR-100) as it requires full backpropagation of the entire networks.

Furthermore, the range of energy consumption is substantial: LifeLearner consumes 0.35\,kJ on GSCv2 while Oracle ANML requires 2.6\,kJ---a 7.4$\times$ difference. Meta-CL methods with the AIM module (ANML+AIM, OML+AIM) consume significantly more energy (e.g., 29.43 kJ for CIFAR-100) compared to more efficient methods like LifeLearner (1.71 kJ) consuming 17.2$\times$ less energy. This increased consumption directly correlates with the computational complexity introduced by the attention mechanisms in AIM, which require additional matrix operations during both forward and backward passes.

For severely resource-constrained devices (like Raspberry Pi Zero 2), LifeLearner offers the near-optimal balance between accuracy and energy efficiency (achieving near oracle accuracy at 2.54-7.43$\times$ lower energy consumption), making it ideal for battery-powered applications with limited computational resources. For devices with moderate resources (Raspberry Pi 3B+), baseline ANML represents a reasonable compromise, providing improved accuracy with only marginal energy increase over pre-trained methods. For devices with greater computational capacity (Jetson Nano), AIM-enhanced methods become viable options when application requirements prioritize accuracy over energy consumption.

\textbf{Generalizability across IoT Hardware.} 
We extended our evaluation to diverse edge devices with varying computational capabilities (Jetson Nano with enhanced specifications and Pi Zero 2 with reduced capacity). The devices we selected span from entry-level edge AI platform (Jetson Nano: 4GB RAM, quad-core processor) to general IoT device (Pi 3B+: 1GB RAM) to highly constrained (Pi Zero 2: 0.5GB RAM) configurations, covering key points in the resource continuum of IoT platforms.
By testing across this hardware spectrum, we observe how different Meta-CL methods perform under varying resource constraints, revealing which approaches are most suitable for specific hardware profiles. Our findings demonstrate that the performance patterns and resource requirements scale predictably with hardware capabilities, allowing practitioners to extrapolate to similar devices. For example, devices with RAM less than 512 MB would likely benefit from the same optimization strategies identified for Pi Zero 2, while those with about 1 GB RAM might follow patterns observed on Pi 3B+. The table~\ref{tab:devices} below shows the end-to-end latency and energy consumption results across these three edge devices, confirming that LifeLearner achieves a 7.5-fold reduction in latency compared to Oracle.

\begin{figure}[t!]
  \centering
  \vspace{-0.2cm}
  \subfloat[UrbanSound8K]{
    \includegraphics[width=0.51\columnwidth]{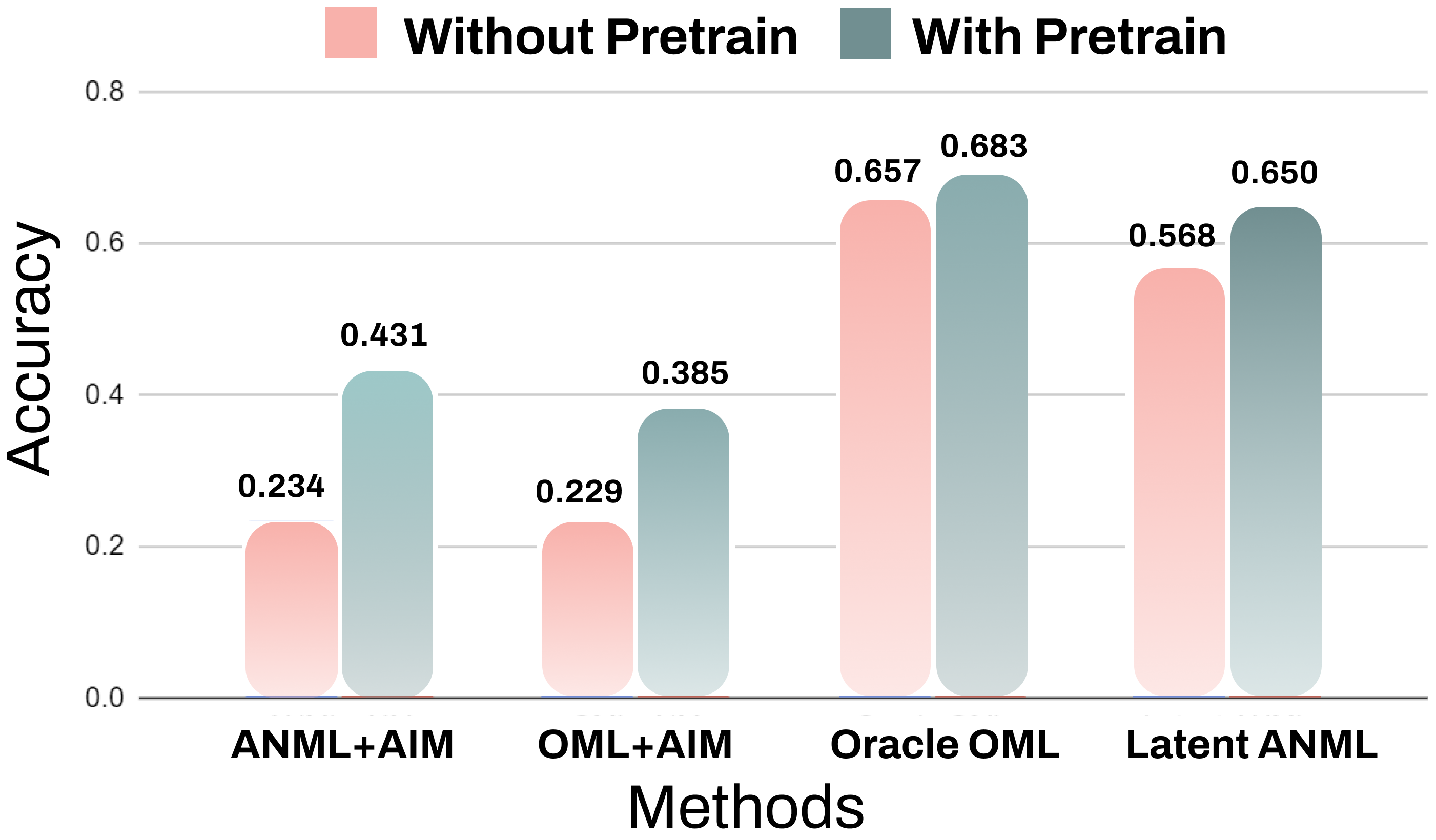}
  \label{subfig:app:pretrain1}}
  \subfloat[ESC-50]{
    \includegraphics[width=0.46\columnwidth]{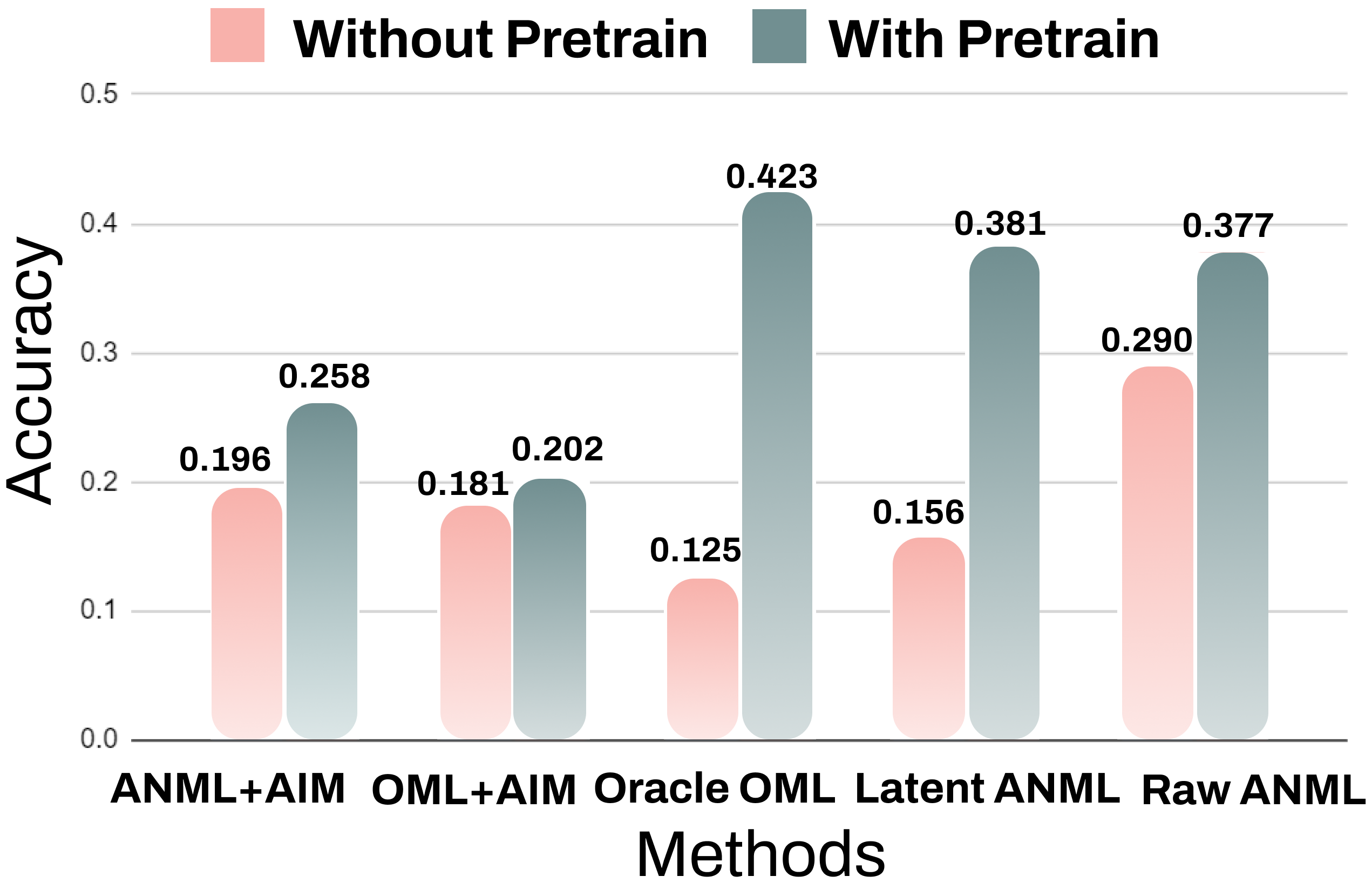}
  \label{subfig:app:pretrain2}}
  \caption{
  Comparison of accuracy with and without pretraining for datasets UrbanSound8K and ESC-50 using the 3-Layer CNN architecture.}
  \label{fig:app:pretraining12}
  \vspace{-0.2cm}
\end{figure}

\begin{table}[t]
  \centering
  \caption{End-to-end latency and energy consumption results of running different Meta-CL methods on three different edge devices}
  \label{tab:devices}
  \resizebox{0.8\columnwidth}{!}{%
  \begin{tabular}{ p {1.4cm} l | c c  }
    \toprule 
    \textbf{Device} &\textbf{Method}  & \textbf{Latency} & \textbf{Energy} \\
    \cmidrule(l){0-3}
     & Pre-trained        & 71s	& 0.32 kJ	\\
    Raspberry        & ANML & 71s & 0.32 kJ  \\
    Pi 3B+       & Oracle ANML & 570s     & 2.60 kJ  \\
              & LifeLearner       & 76s	& 0.35 kJ	\\
    \cmidrule(l){0-3}
     & Pre-trained        & 79s	& 0.36 kJ	\\
    Jetson    & ANML        & 79s	& 0.36 kJ	\\
    Nano     & Oracle ANML & 633s & 2.00 kJ  \\
              & LifeLearner & 84s     & 0.39 kJ  \\
    \cmidrule(l){0-3}
     & Pre-trained        & 74s	& 0.33 kJ	\\
    Raspberry        & ANML & 75s & 0.33 kJ  \\
    Pi Zero 2    & Oracle ANML & 601s     & 2.64 kJ  \\
    & LifeLearner & 80s    & 0.35 kJ  \\
    \bottomrule
  \end{tabular}
  }
\end{table}

\subsection{Analysis and Discussion}\label{subsec:analysis}

\textbf{Effectiveness of Pre-training.} 
We examine the effectiveness of pre-training on the performance of Meta-CL methods.
Our evaluation across five datasets without pre-training reveals varying performance levels: while the two image datasets and GSCv2 demonstrated satisfactory accuracy, UrbanSound8K and ESC-50 showed suboptimal performance. Following the approach used in advanced models, we implemented pre-training before meta-training for each method. As shown in Figure~\ref{fig:app:pretraining12}, the incorporation of pre-training significantly enhanced model performance. Notably, Meta-CL methods achieved an average performance improvement of 13.98\% on the UrbanSound8K and ESC-50 datasets after pre-training. Our analysis revealed that CNNs exhibited considerable performance degradation when meta-training was conducted without pre-training. Pre-training plays a significant role in enhancing accuracy, particularly for ANML and OML, where accuracy without pre-training is a mere 5\%. Advanced architectures such as ViT and YAMNet intrinsically incorporate pre-training, obviating the need for ablation study of pre-training with them.

\begin{figure}[t!]
  \vspace{-0.2cm}
  \centering
  \subfloat[Oracle \& Latent (YAMNet)]{
    \includegraphics[width=0.485\columnwidth]{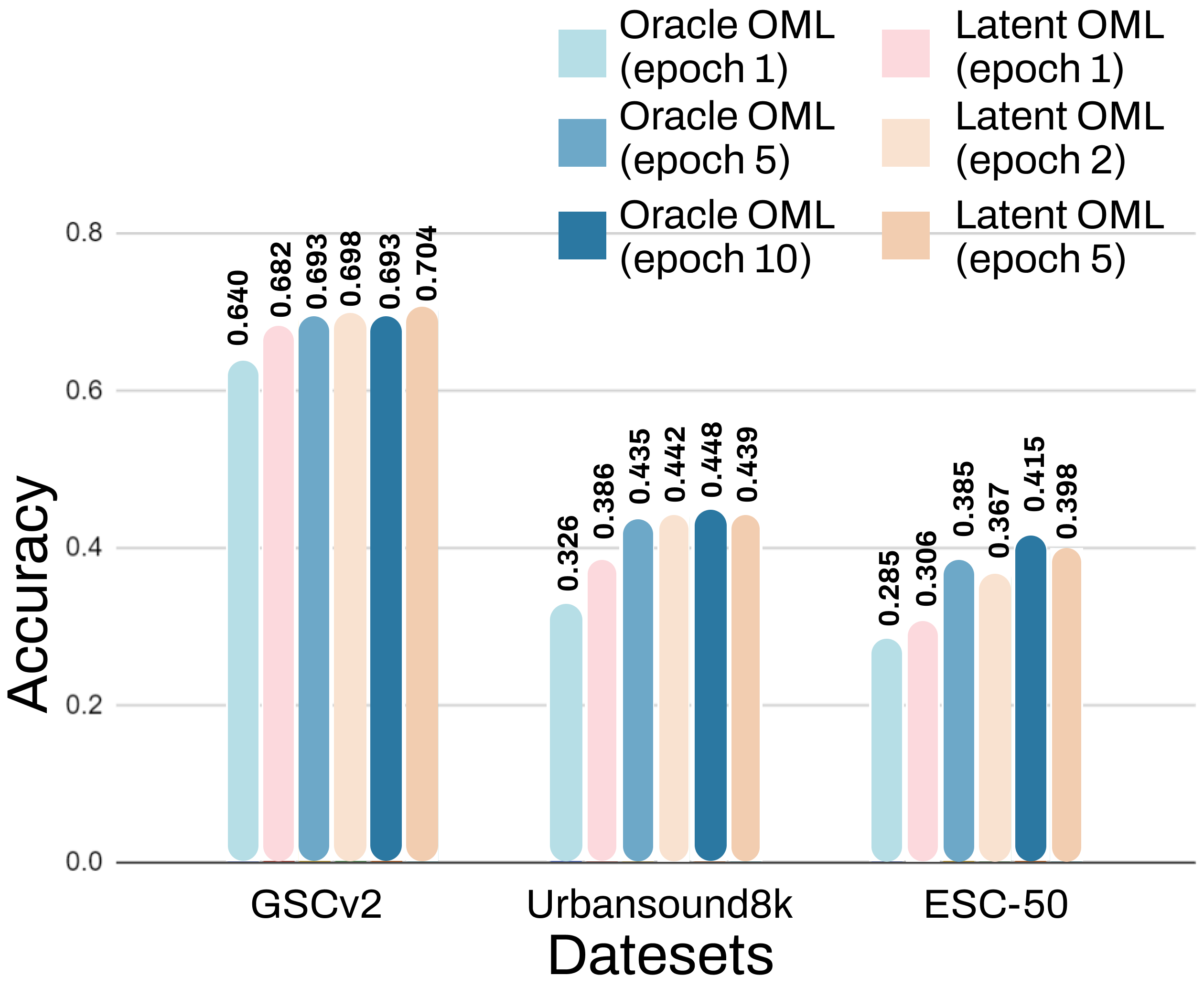}
  \label{subfig:app:epoch11}}
  \subfloat[Raw \& Latent (YAMNet)]{
    \includegraphics[width=0.485\columnwidth]{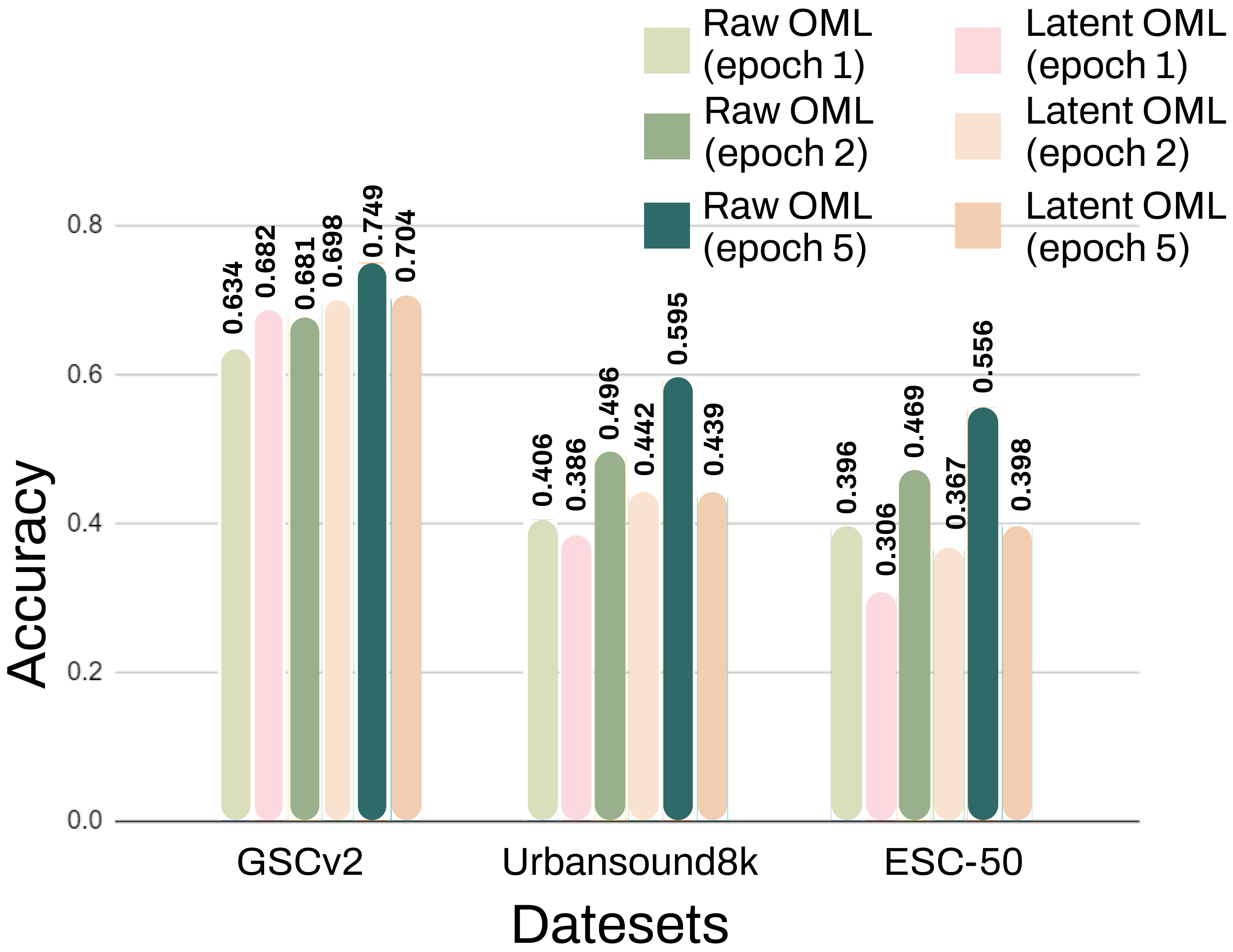}
  \label{subfig:app:epoch21}}
  \hfill
  \centering
  \subfloat[Oracle \& Latent (CNN)]{
    \includegraphics[width=0.485\columnwidth]{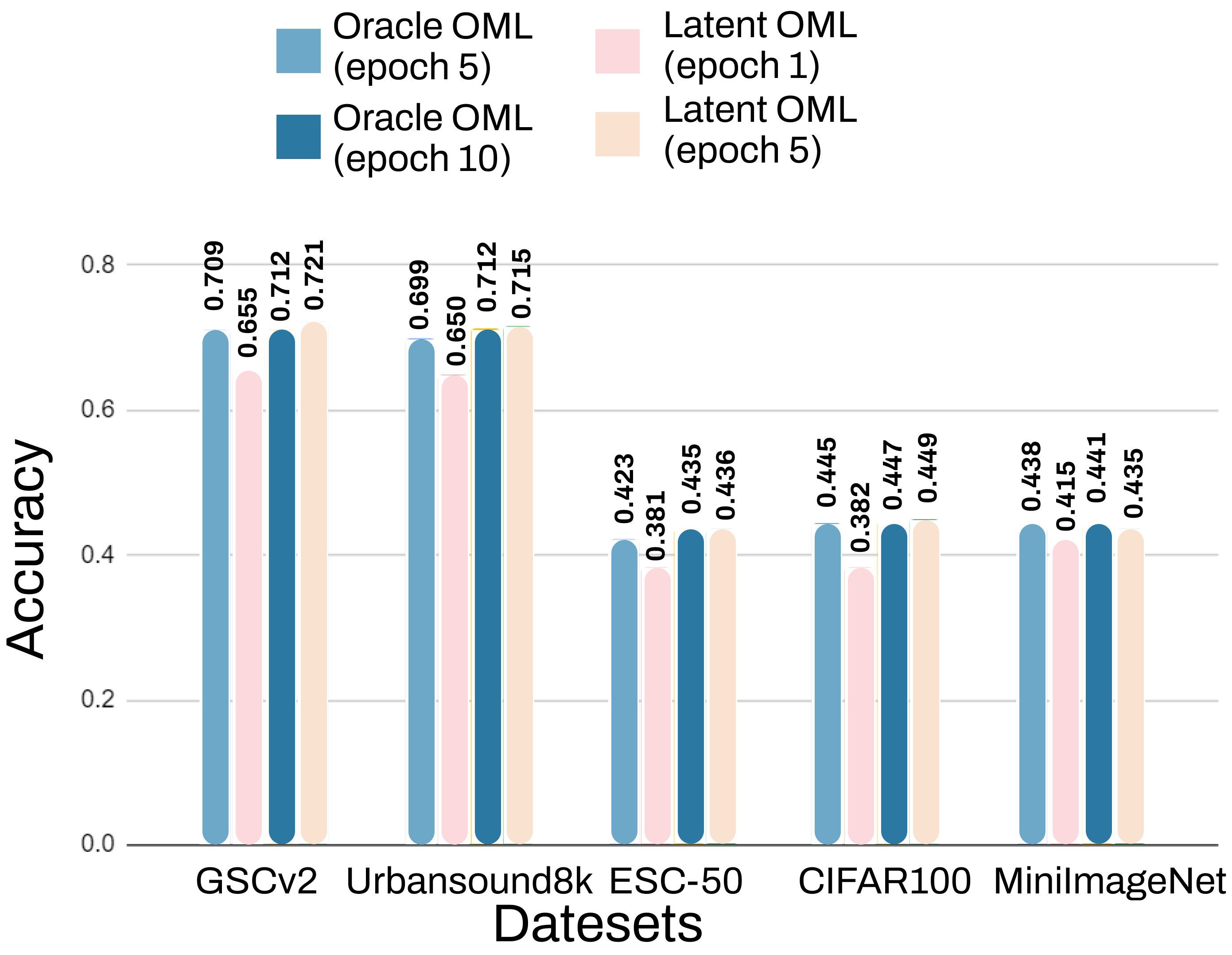}
  \label{subfig:app:epoch41}}
  \subfloat[Raw \& Oracle \& Latent (ViT)]{
    \includegraphics[width=0.485\columnwidth]{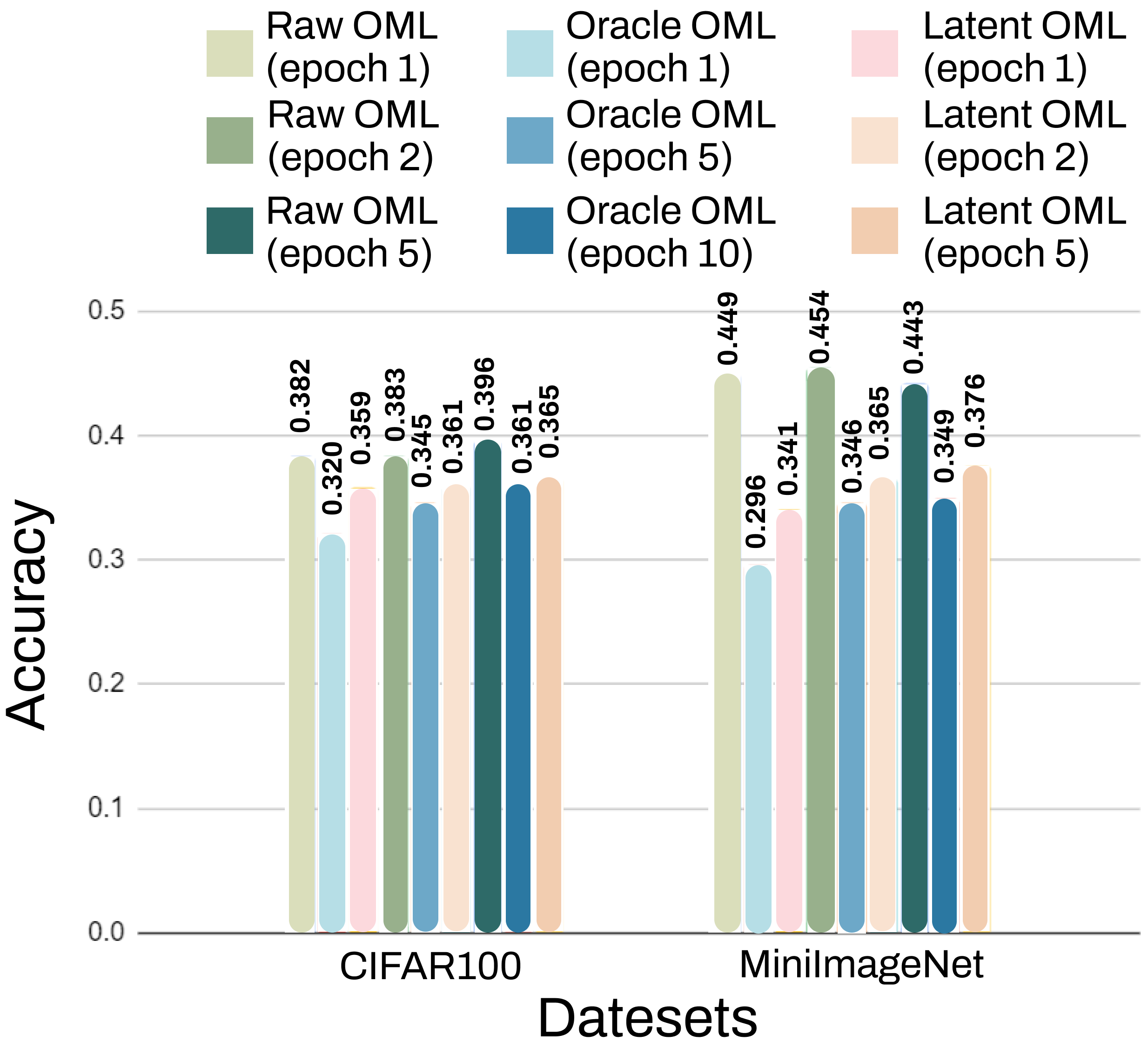}
  \label{subfig:app:epoch31}}
  \caption{
  Comparison of different Meta-CL methods regarding the number of replay epochs.
  }
  \label{fig:app:epoch1234}
  \vspace{-0.2cm}
\end{figure}

\textbf{The Number of Replay Epochs.}
We investigated the relationship between replay frequency and accuracy, acknowledging that increased replays correspond to higher latency and energy consumption. Recognizing that an increased number of replays correlates with heightened latency and energy usage, we explored the association between replay frequency and accuracy. Figure ~\ref{subfig:app:epoch11} indicates that Oracle OML necessitates more than five replays to attain elevated accuracy levels, thereby escalating energy expenditure throughout the training phase. Conversely, Latent OML demonstrates superior performance, requiring merely one or two replays to reach comparable accuracy, thereby diminishing system overhead. In the case of the same dataset, the efficiency of Latent OML after five replays approximates that of Oracle OML following ten replays. The same observation is evident in the CNN architecture depicted in Figure ~\ref{subfig:app:epoch41}; here, one or two iterations of the Latent method attain the same level of accuracy as the Oracle method after five iterations. Additionally, we note that exceeding five iterations does not significantly enhance accuracy. Figures~\ref{subfig:app:epoch21} and~\ref{subfig:app:epoch31} illustrate that the accuracy of Raw OML within the advanced methods surpasses that of the previously tested Latent approach.

\textbf{Analysis of AIM and Compression Module.}
In CNN architectures, ANML with AIM demonstrates substantial accuracy improvements (from 26.0\% to 34.6\% on CIFAR-100), albeit at the cost of increased memory requirements (from 39.69\,MB to 1,093\,MB). The integration of AIM, while beneficial for accuracy, introduces significant computational overhead. LifeLearner presents an optimal balance between accuracy and memory utilization in resource-constrained IoT devices via efficient compression module for latent replays.

Vision Transformer (ViT) and YAMNet architectures exhibit analogous performance-resource tradeoffs. ViT implementations with OML and AIM show improved accuracy but incur substantial computational costs (22,968s latency, 103 kJ energy consumption on CIFAR-100). Similarly, YAMNet with AIM achieves superior accuracy (from 42.9\% to 71.0\% on GSCv2) while requiring increased memory utilization (from 10.16\,MB to 608.2\,MB). Pre-trained YAMNet models demonstrate superior resource efficiency metrics.

\subsection{Differences  between image and audio datasets}\label{subsec:difference}

IoT deployments span both image sensors (cameras, visual inspection systems) and audio sensors (microphones for voice interfaces, acoustic monitoring). Understanding modality-specific Meta-CL behavior is essential for system design.

First, the temporal nature of audio data creates distinct feature dependencies compared to image data. Our experimental results demonstrate that ANML achieves 59.6\% accuracy on UrbanSound8K versus only 32.7\% on MiniImageNet, a disparity stemming from audio's sequential characteristics being better suited for neural networks to extract temporal patterns. 

Also, audio data requires specialized preprocessing steps (e.g., spectrogram conversion or MFCC feature extraction) that introduce extra complexity in Meta-CL systems. Architectures optimized for audio like YAMNet exhibit different memory-energy tradeoffs compared to visual architectures like ViT, with the former showing better energy efficiency on audio samples (5.4 kJ for UrbanSound8K vs. 11.2 kJ for MiniImageNet).

Finally, we find that audio Meta-CL is more sensitive during the pre-training phase, as in Figure~\ref{fig:app:pretraining12}, where pre-training improved average performance by 13.98\% for UrbanSound8K and ESC-50 datasets, while the improvement for image datasets was substantially smaller. This indicates that knowledge transfer in the audio domain requires stronger domain-specific foundations to work effectively in Meta-CL settings.

\vspace{-1mm}
\section{Practical Guidelines}\label{sec:guideline}

Our extensive empirical analysis provides evidence-based guidelines for researchers and practitioners implementing Meta-CL on edge devices.
While our investigation focuses on specific Meta-CL methods and may not encompass all learning approaches, our results and findings show consistent patterns across different architectures and datasets, which lead to the following key guidelines.

\textbf{Resource-Aware Method Selection.} 
Device memory during runtime (not nominal RAM) is the primary constraint determining method viability. For devices with less than 1 GB available RAM, AIM-enhanced methods should be excluded due to OOM failures observed in our experiments. LifeLearner and Latent OML provide the best balance of accuracy and resource efficiency across memory-constrained environments. For devices with sufficient memory ($\geq$2 GB available), AIM-enhanced methods become viable when accuracy is paramount.

\textbf{Architecture Selection.} 
Direct comparison across architectures is limited because CNN uses ANML-based methods while ViT and YAMNet use OML-based methods. Nevertheless, we observe that advanced architectures (ViT and YAMNet) do not necessarily guarantee better Meta-CL performance over CNN. For image tasks, results are dataset-dependent: CNN with Oracle ANML outperforms ViT with Oracle OML on CIFAR-100 (44.5\% vs. 36.1\%) but underperforms on MiniImageNet (43.8\% vs. 51.2\%). For audio tasks, YAMNet tends to achieve higher accuracy than CNN while requiring less memory. These findings challenge conventional assumptions about model complexity.

\textbf{Pre-training Strategy.} 
Pre-training before meta-training is essential, particularly for audio domains where we observed 14\% average accuracy improvement (Figure~\ref{fig:app:pretraining12}). As detailed in Section~\ref{subsec:Experimental Setup}, for CNNs, we pre-train on source classes until the validation loss converges. For YAMNet and ViT, we leverage publicly available pre-trained weights directly.

\textbf{Edge Deployment Considerations.} Latent methods such as LifeLearner for CNN and Latent OML for ViT/YAMNet consistently demonstrate strong performance across image and audio datasets, balancing accuracy with resource efficiency. These methods are particularly advantageous for battery-powered IoT applications, enabling real-time category integration, efficient personalised processing, and privacy-preserving local computation without additional communication overhead.

AIM-enhanced methods should be avoided in energy-constrained scenarios due to their substantial memory and high energy usage. While Raw ANML and Raw OML can sometimes achieve higher accuracy than Oracle methods, their extreme latency makes them impractical for edge deployment. For example, on CIFAR-100, Raw ANML and Raw OML require 11,424s and 22,968s, respectively, compared to 374s and 764s for LifeLearner and Latent OML -- over 30$\times$ slower.

\vspace{-1mm}
\section{Conclusion}\label{sec:conclusions}

In this paper, we presented \sysname, a benchmark framework that evaluates Meta-CL methods for IoT deployment viability, not just accuracy. Our evaluation across six methods, three architectures, five datasets, and three device classes reveals that deployment constraints significantly narrow method choices: several methods fail on sub-1 GB devices, LifeLearner achieves near-oracle accuracy at 2.54-7.43$\times$ lower energy consumption. Pre-training followed by meta-training improves accuracy, and lightweight CNNs outperform larger architectures under few-shot constraints.

\textbf{Limitations \& Future Work.} Our findings may not generalise to platforms with neural accelerators (e.g., Google Coral) or microcontrollers with extremely limited memory (less than 1 MB). Future work should investigate long-term deployment stability.

\bibliographystyle{IEEEtran}
\bibliography{main}

\end{document}